\documentclass[10pt,twocolumn,letterpaper,final]{article}
\usepackage{wacv}
\usepackage{booktabs} 
\usepackage{framed,multirow}
\usepackage{amssymb}
\usepackage{latexsym}
\usepackage{graphicx}
\usepackage{amsmath}
\usepackage{array}
\usepackage{caption}
\usepackage{subcaption}
\usepackage{url}
\usepackage{xcolor}
\usepackage{pifont}
\date{}

\definecolor{OliveGreen}{rgb}{0,0.6,0}
\definecolor{mustard}{rgb}{1,0.79,0.04}
\newcommand*\colourcheck[1]{%
  \expandafter\newcommand\csname #1check\endcsname{\textcolor{#1}{\ding{52}}}%
}
\newcommand{\cmark}{\ding{51}}%
\newcommand{\xmark}{\ding{55}}%
\colourcheck{blue}
\colourcheck{green}
\colourcheck{red}

\wacvfinalcopy 

\begin{document}

\title{WSD: Wild Selfie Dataset for Face Recognition in Selfie Images}

\author{Laxman Kumarapu$^{1,*}$, Shiv Ram Dubey$^{2,*}$, Snehasis Mukherjee$^{3,*}$, Parkhi Mohan$^{4,*}$, \\ Sree Pragna Vinnakoti$^{5,*}$, Subhash Karthikeya$^6$\\
\textit{$^1$New York University, USA}\\
\textit{$^2$CVBL, Indian Institute of Information Technology, Allahabad, India}\\
\textit{$^3$Shiv Nadar University, India}\\
\textit{$^4$Indian Institute of Management, Lucknow, India}\\
\textit{$^5$Arizona State University, USA}\\
\textit{$^6$Indian Institute of Information Technology, Sri City, Chittoor, India}\\
{\small Email: srdubey@iiita.ac.in}
}

\maketitle







\begin{abstract}
\let\thefootnote\relax\footnote{$^*$Authors were affiliated to IIIT Sri City at the time of creation of dataset.}
With the rise of handy smart phones in the recent years, the trend of capturing selfie images is observed. Hence efficient approaches are required to be developed for recognising faces in selfie images. Due to the short distance between the camera and face in selfie images, and the different visual effects offered by the selfie apps, face recognition becomes more challenging with existing approaches. A dataset is needed to be developed to encourage the study to recognize faces in selfie images. In order to alleviate this problem and to facilitate the research on selfie face images, we develop a challenging Wild Selfie Dataset (WSD) where the images are captured from the selfie cameras of different smart phones, unlike existing datasets where most of the images are captured in controlled environment. The WSD dataset contains 45,424 images from 42 individuals (i.e., 24 female and 18 male subjects), which are divided into 40,862 training and 4,562 test images. The average number of images per subject is 1,082 with minimum and maximum number of images for any subject are 518 and 2,634, respectively. The proposed dataset consists of several challenges, including but not limited to augmented reality filtering, mirrored images, occlusion, illumination, scale, expressions, view-point, aspect ratio, blur, partial faces, rotation, and alignment. We compare the proposed dataset with existing benchmark datasets in terms of different characteristics. The complexity of WSD dataset is also observed experimentally, where the performance of the existing state-of-the-art face recognition methods is poor on WSD dataset, compared to the existing datasets. Hence, the proposed WSD dataset opens up new challenges in the area of face recognition and can be beneficial to the community to study the specific challenges related to selfie images and develop improved methods for face recognition in selfie images.
\end{abstract}


\section{Introduction}
Face recognition has been a challenging problem for computer vision researchers during the past few years \cite{zhao2003face}, \cite{abate20072d}. Earlier hand-crafted feature representation and classical machine learning based methods were utilized for face recognition and analysis \cite{saber1998frontal}, \cite{wang2002facial}, \cite{deniz2003face}, \cite{ahonen2004face}, \cite{he2005face}, \cite{deniz2011face}, \cite{ldop}, \cite{chakraborty2018centre}, \cite{fdlbp}. Such hand-crafted feature-based methods often used to be influenced by the visual characteristics of the image, such as skin, texture, shape and facial parts features. These methods relied on traditional machine learning tools such as K-Nearest Neighbor (KNN), Support Vector Machine (SVM), Decision Tree, Random Forest, etc. for recognition.

After 2010, deep neural networks (DNNs) have emerged as a powerful learning mechanism in solving various tasks in computer vision, including face recognition. DNNs can learn the important features automatically from the image data in a hierarchical fashion at different levels of abstractness \cite{lecun2015deep}.

The deep learning mechanism has shown a huge improvement in 2012 for image recognition using convolutional neural networks (CNNs) \cite{alexnet}. Since then several variants of CNNs have been proposed for different applications of computer vision and image processing \cite{vgg}, \cite{resnet}, \cite{fasterrcnn}, \cite{imageretrieval}, \cite{feng2022iris}. The key success of deep learning models was observed due to the availability of powerful computational machines, such as Graphical Processing Units (GPUs) which can perform heavy parallel computation, and availability of large-scale datasets \cite{imagenet}.

The recent advancements in face recognition research, with the help of deep learning methods, can be observed from the outstanding performance of deep learning models \cite{srivastava2019performance}, such as VGGFace \cite{vggface}, \cite{vggface2} and FaceNet \cite{facenet}. So far, several deep learning models have been developed by the researchers for face recognition by introducing different CNNs, different objective functions, and different other strategies \cite{deepface}, \cite{sphereface}, \cite{arcface}, \cite{hmloss}, \cite{mishra2021multiscale}. A detailed discussion on the advancements of deep learning based face recogntion models can be found in a recent survey published by Wang and Deng in \cite{wang2021deep}.

The recent progress in deep learning based face recognition technology is heavily governed by the large-scale face datasets. Hence, several benchmark face datasets have been developed to drive the research and development activities in face recognition. The popular face datasets include Labelled Faces in Wild \cite{lfw}, CASIA WebFace \cite{casiawebface}, VGGFace \cite{vggface}, VGGFace2 \cite{vggface2}, UMDFace \cite{umdface}, Microsoft Celeb (MS-Celeb-1M) \cite{msceleb} and WebFace \cite{webface}. We present a detailed comparison of the different face datasets in terms of the size, annotations and different characteristics in Table \ref{tab:challengeComparison} and Table \ref{tab:datasetComparison}. Some other face datasets include multi-face dataset \cite{dubey2018multi}, YouTube Face Dataset \cite{wolf2011face}, CelebFaces Attributes Dataset (CelebA) \cite{liu2015faceattributes}, etc. These existing datasets were generally developed by mining the images of celebrities from the public domain. The major focus of these datasets is on more and more samples. Hence, some of these datasets are also noisy and does not contain the scenarios where users take their pictures by themselves using their smart phones, i.e., selfie images.

Some reasearchers have tried to come across the datasets which include the selfie images, such as Selfie Dataset\footnote{https://www.crcv.ucf.edu/data/Selfie/} by Kalayeh et al. \cite{kalayeh2015take}, DocFace \cite{shi2018docface} \& DocFace+ \cite{shi2019docface+} by Shi and Jan, and SelfieCity\footnote{https://selfiecity.net/}. The images in Selfie Dataset \cite{kalayeh2015take} is collected from Instagram and annotated with different attributes. Hence the Selfie Dataset \cite{kalayeh2015take} is used for analyzing ``How to Take a Good Selfie" instead of face recognition. However, we aim to develop a wild selfie dataset, which poses a great challenge by avoiding any environmental constrains while taking the selfies. Further, the proposed dataset should be useful for face recognition in selfie images. The DocFace performs the matching of ID photos with live face photos/selfies \cite{shi2018docface}. The DocFace+ supports multiple ID/selfie per class for ID-Selfie matching \cite{shi2019docface+}.
The SelfieCity dataset is collected from Instagram and labelled based on the Cities. This dataset is also not available publicly for the research purposes. 

The existing selfie datasets are either not available publicly or not suitable for face recognition in selfie images. Thus, through this paper, we propose a selfie dataset for face recognition. Following are the contributions of this paper:
\begin{enumerate}
    \item A wild selfie dataset (WSD) containing more than 45K images from 42 individuals is proposed to facilitate face recognition in selfie images captured using smartphone cameras.
    \item The proposed dataset poses a great challenge in terms of severe variability in image characteristics, such as aspect ratio, partial faces, occlusion, illumination, scale, expression, viewpoint, blur, reflection and augmented reality.
    \item We test the performance of existing deep learning based face recognition models to show the complexity of the proposed WSD dataset.
\end{enumerate}

The rest of the paper is presented by detailing the proposed dataset in Section 2, experimental setup in Section 3, results in Section 4, and conclusion in Section 5.

\begin{figure*}[!t]
    \centering
    \includegraphics[width=\textwidth,height=15cm]{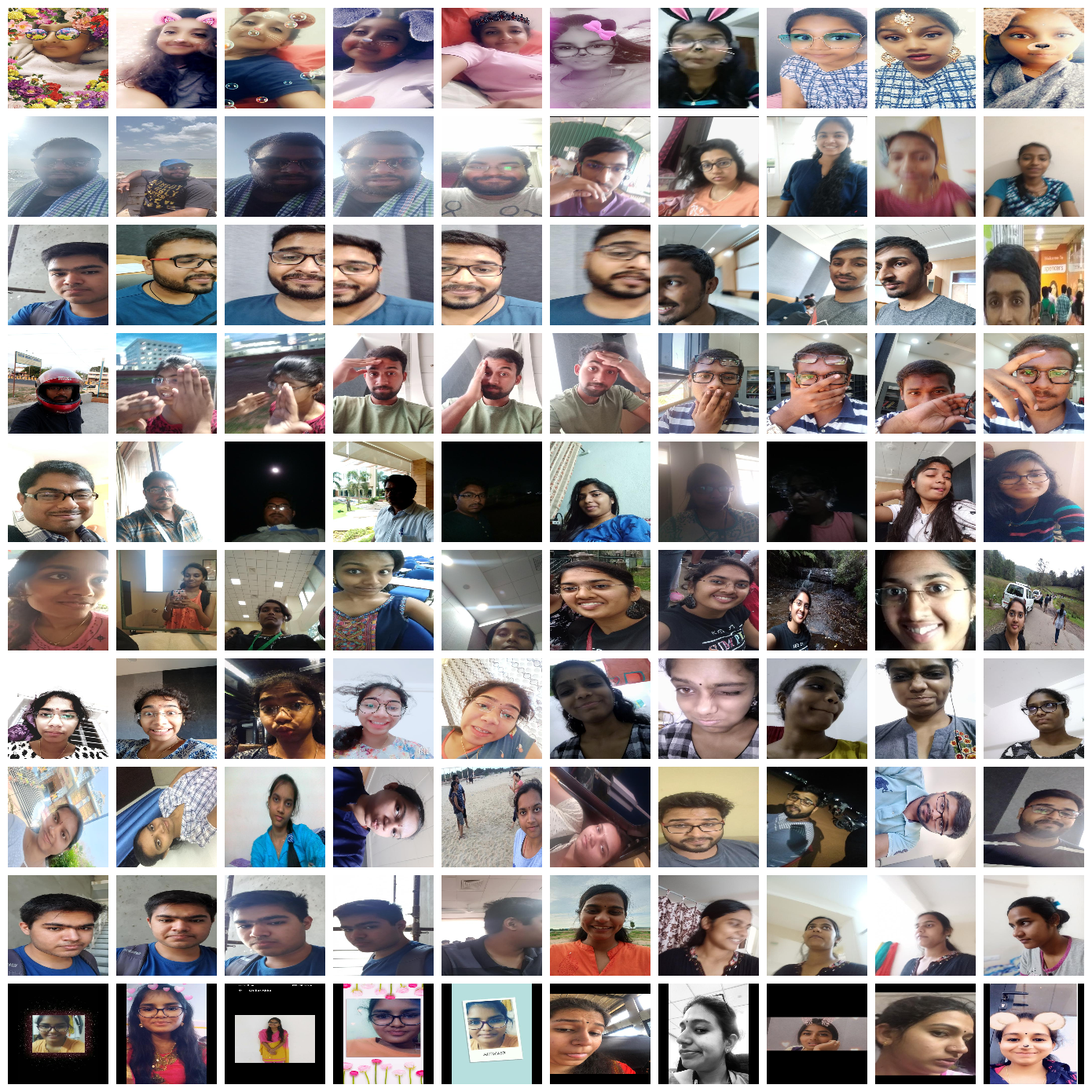}
    \caption{Sample images depicting the challenges in the proposed WSD dataset. The rows from top to bottom include the images having effect of AR Filters, Mirrored Reflections, Blurred Images, Partial Faces, Occlusions, Illumination Changes, Scaling, Different Expressions and Emotions, Different Alignments, View Point Variation and Different Aspect Ratios.}
    \label{fig:wsdExamples}
\end{figure*}

\section{Proposed Wild Selfie Dataset}
\begin{figure*}[!t]
    \centering
    \includegraphics[trim={7mm 7mm 7mm 20mm},clip, width=\columnwidth]{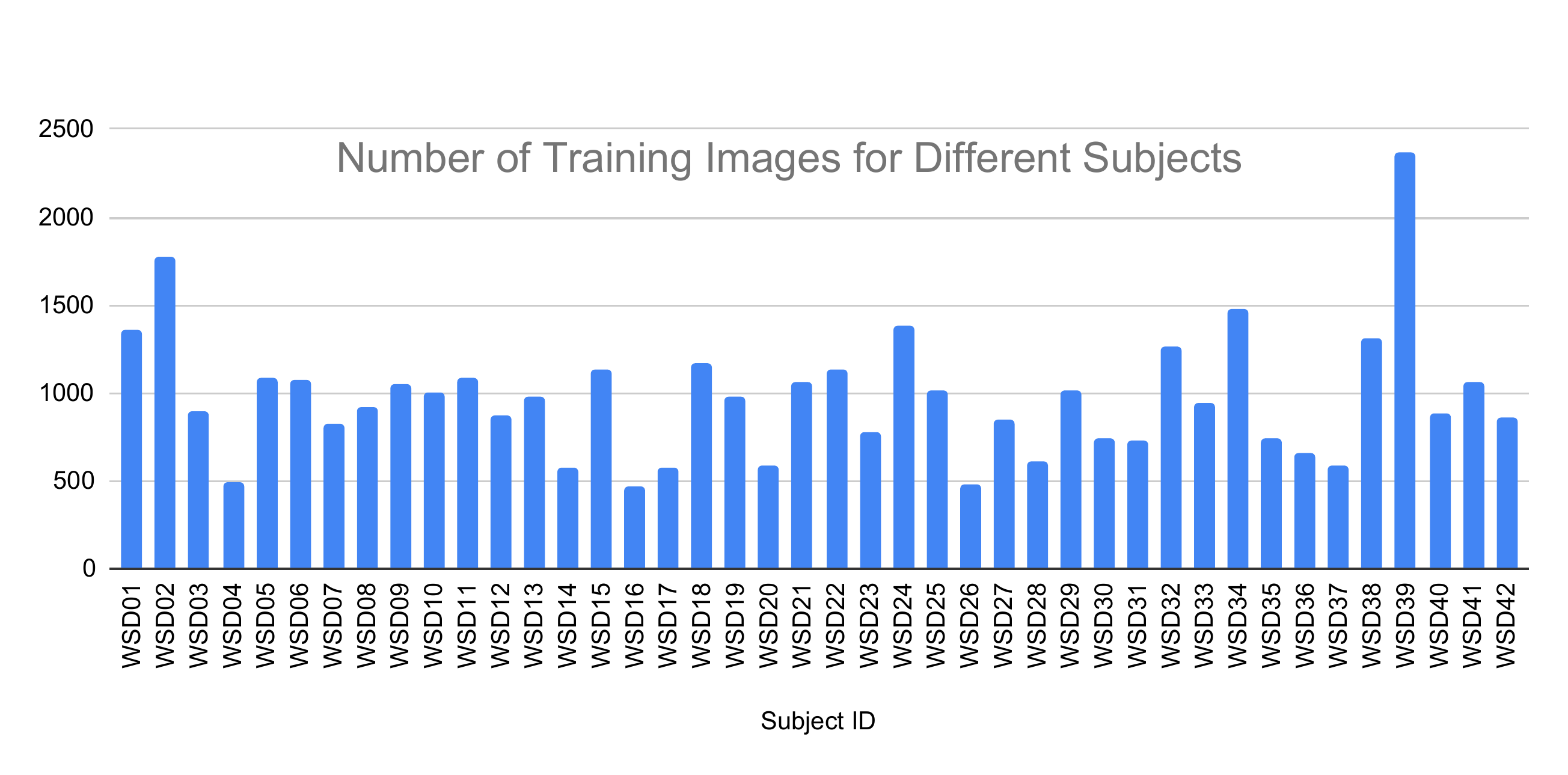}
    \hspace{3mm}
    \includegraphics[trim={7mm 7mm 7mm 20mm},clip, width=\columnwidth]{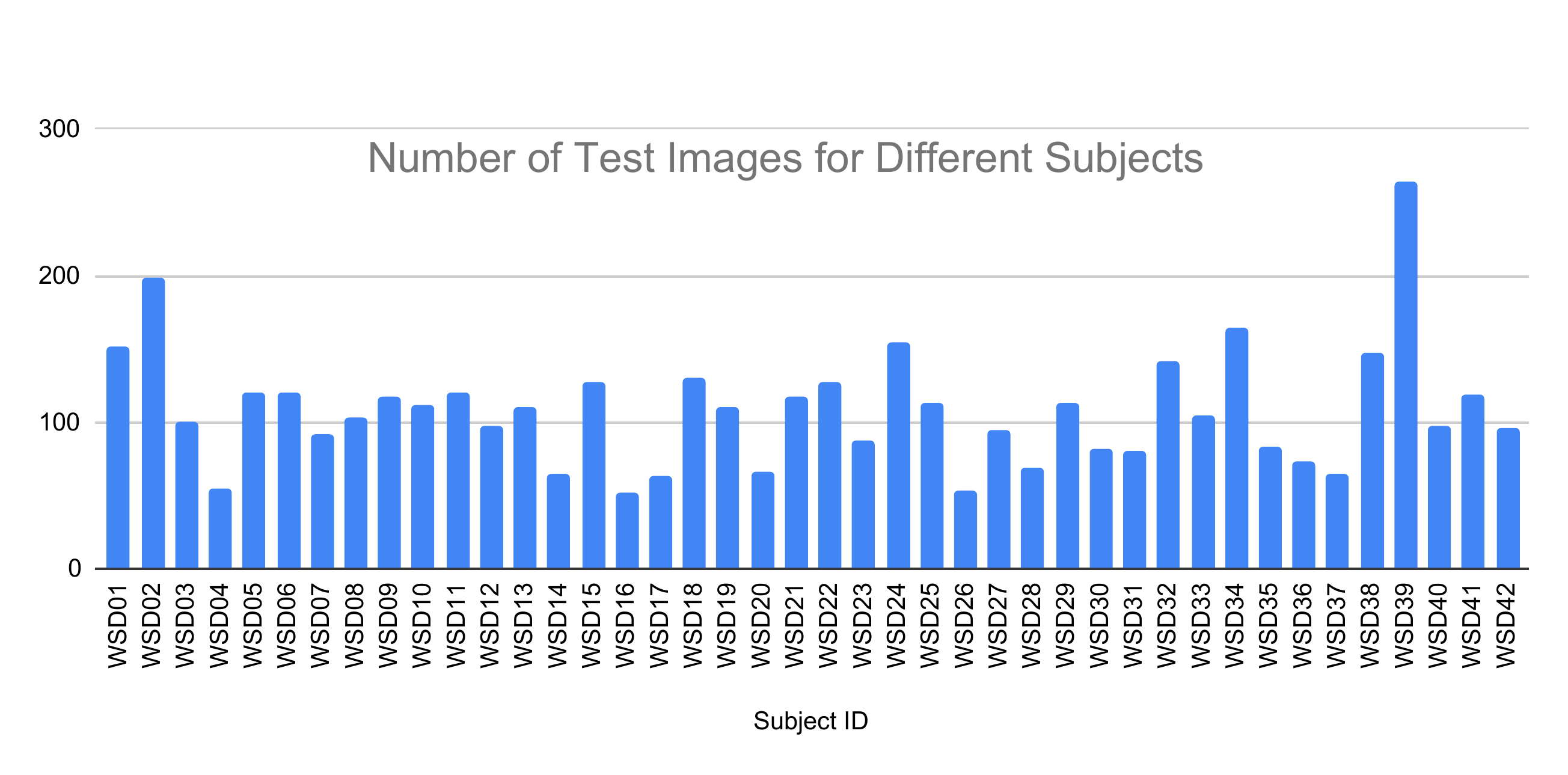}
    \caption{The number of samples in different categories of WSD dataset under training and test sets.}
    \label{fig:train-test}
\end{figure*}

This section describes the proposed Wild Selfie Dataset (WSD) including the selfie image collection process, dataset characteristics, different challenges and a comparison with the existing face datasets. The challenges in WSD include combinations of diverse and difficult real-world selfie image scenarios that humans click selfies in, like augmented reality (AR) filters, blurred and occluded images, mirrored reflections, etc. The sample images depicting different challenges in the dataset are shown in Fig. \ref{fig:wsdExamples}. The dataset comprises of a total of 45,424 selfie images from 42 participants including 24 females and 18 males of 18-31 years age group. The average number of samples in each category is 1,082 with minimum and maximum number of samples in a category as 518 and 2,634, respectively. The labelling is performed manually in terms of the bounding box annotations, subject class levels, and gender of subjects. The dataset is divided into training and test sets with 40,862 and 4,562 images, respectively. The category-wise number of training and test images is summarized in Fig. \ref{fig:train-test}. It can be observed that the dataset is splitted in the training and test sets uniformly.

\subsection{WSD Dataset Creation}
The procedure of creation of the proposed WSD Dataset consists of the following steps: Raw data collection, Pre-processing raw data, Near duplicate elimination, Face detection and bounding box annotation, Manual Filtering and Human Annotation, and Labelling the faces. Next we describe each step in detail.

\subsubsection{Raw Data Collection}
Participants were asked to contribute to the Wild Selfie Dataset by submitting selfie images and self-recorded videos. A selfie is a photograph that a person takes of himself/ herself. Selfie images may be captured through the front or rear camera of the phone, with the aid of a selfie stick, or using laptop cameras, provided the captured images include the human face. The selfie videos were strictly restricted to be taken through the front camera of a smartphone which captured the face of the subject. The agreement has been signed with the participants to use their images for non-commercial research and development purposes.

\subsubsection{Pre-Processing Raw Data}
After collection of the raw data of selfie images and videos, the unsupported file formats and corrupted images are removed from the dataset. Then, the images are extracted from video frames. The multimedia framework FFmpeg\footnote{FFmpeg Developers: http://ffmpeg.org/} is used to extract images from each video. For each individual video, different number of images are extracted depending on expression, illumination, surrounding background and other variations.

\subsubsection{Near Duplicate Elimination}
After extracting images from video frames, all data are in the form of images. At this stage duplicate elimination is imperative so when a model is trained on the dataset, it is not affected by undue bias. A duplicate image in the Wild Selfie Dataset collection is defined as one which is an exact (duplicate) or almost exact (near duplicate) pixel-to-pixel match to an identical image or is obtained from another image through cropping or rescaling techniques. An earnest attempt to remove duplicates in the WSD has been made by us. There exist image pairs in the final WSD dataset, that are similar, but not the exact same. A clear distinction is visible in variations that may be present in the face position, orientation, illumination or expression of the person.

\subsubsection{Face Detection and Bounding Box Annotation}
Following near duplicate image removal, unique selfies of each participant are obtained. The annotation of the WSD Dataset is performed by generating the face bounding boxes. The Dlib \cite{dlib09} general-purpose cross-platform software library is used to detect the faces and to obtain the top-left and bottom-right facial bounding box coordinates. The coordinates are used to compute the width and height of the bounding box. Thus, the final bounding box annotations include the top-left coordinates (X, Y), the width (W), and the height (H).
Upon human evaluation of the generated bounding boxes, we find the following issues: 1) not all bounding boxes contain faces, 2) some bounding boxes include the faces having severe occlusion with others, and 3) some faces remain undetected by Dlib. Thus, manual checking and annotations are also performed, as described next, to correct the above issues.

\subsubsection{Manual Filtering and Human Annotation}
As mentioned above, the manual checking and human annotation are performed to fix the discrepancies caused by Dlib face detection. For each image, the bounding boxes generated through Dlib are manually verified and corrected using the following three criteria: 1) if the image contains a bounding box having a face of the corresponding person then the same is retained, 2) if the image contains a bounding box having a face of a person other than the labeled person, then that bounding box is removed and the bounding box of the corresponding individual is then manually computed, and 3) if Dlib is unable to detect a face in the selfie then the bounding box of the corresponding participant is manually computed.

\subsubsection{Labelling the Faces}
We also provide the face identities of the subjects, for use in face recognition task. In order to make the labels independent of actutal identity of the persons involved, we provide the face identities using the unique codes, such as WSDXX, where XX is from 01 to 42 for all 42 persons participated in data collection. The gender of all participants are also labelled.

\begin{figure*}[!t]
    \centering
    \includegraphics[width=0.328\textwidth]{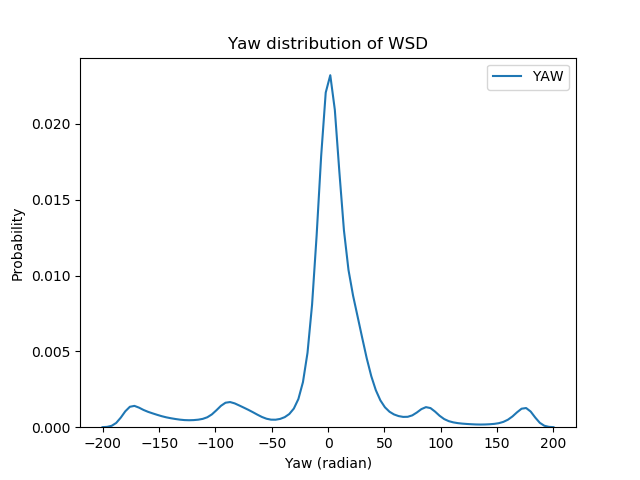}
    \includegraphics[width=0.328\textwidth]{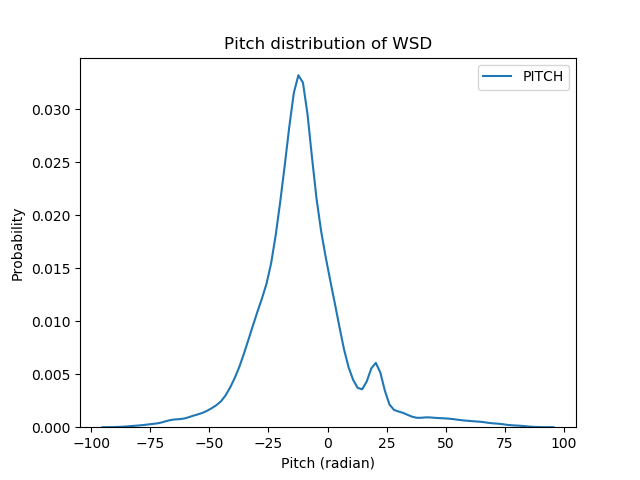}
    \includegraphics[width=0.328\textwidth]{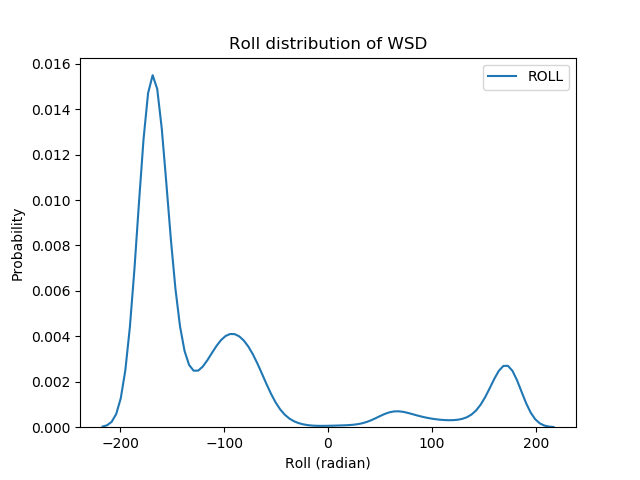}
    \caption{Histogram of Yaw angles (left), Pitch angles (middle), and Roll angles (right) of faces in the WSD dataset.}
    \label{fig:yaw-pitch-roll}
\end{figure*}

\subsection{Dataset Distribution}
To analyse the camera movement, the distribution of the WSD dataset is calculated through head pose estimation. The head pose is determined by the position of the human face and it's allignment. A camera can be rotated in three ways: side-to-side, up-and-down, and around the optical axis with three corresponding angles, i.e., Yaw, Pitch, and Roll, respectively. OpenCV tool \cite{opencv_library} and 6 key landmark points are used to estimate the 3D head pose of images. The six landmarks used are: Left-Eye-Left-Corner, Right-Eye-Right-Corner, Mouth-Left-Corner, Mouth-Right-Corner, Nose-Center-Tip, and Chin-Center-Tip. Yaw angles vary around the perpendicular axis. Yaw positive values depict the right side positions and negative values depict the left side positions. Pitch angles vary across the lateral axis. Pitch positive values depict the upward direction and negative values depict the downward direction. Roll angles vary around the longitudinal axis. Roll positive values depict clockwise rotation and negative values depict counter-clockwise rotation. The yaw, pitch and roll distributions of the proposed WSD dataset are illustrated in Fig. \ref{fig:yaw-pitch-roll}. The yaw distribution shows that more people are inclined towards a slight right profile, though variation is throughout visible. The pitch distribution shows that people tilt their head more towards the bottom in comparison to top while clicking the pictures. The roll distribution shows that the left head rotation is present heavily than the right counterpart. It is also interesting to note that no image head is exactly straight.

\subsection{WSD Dataset Characteristics and Challenges}
\indent The proposed WSD dataset poses multiple not-available-before challenges in a single dataset that mimic real-world selfie data. The major challenges included in WSD dataset are AR (Augmented Reality) filters, Mirrored reflections, Blurred images, Partial faces, Occlusions, Illumination changes, Scaling variations, Different expressions and emotions, Variation in alignments, View Point variation, and Different aspect ratios. The following subsections describe in detail on how the WSD dataset defines and incorporates the challenges mentioned above.

\begin{table*}[!t]
\centering
\caption{Comparison of challenges in the proposed WSD dataset vs the existing face datasets. Here, \textcolor{OliveGreen}{\cmark}: Present, \textcolor{red}{\xmark}: Absent, and \textcolor{mustard}{\textbf{P}}: Partial.}
\begin{tabular}{ | m{0.17\textwidth} | m{0.05\textwidth} | m{0.06\textwidth} | m{0.08\textwidth} | m{0.06\textwidth} | m{0.06\textwidth} | m{0.06\textwidth} | m{0.047\textwidth} | m{0.048\textwidth} | m{0.06\textwidth} | m{0.04\textwidth} |}
 \hline
 \multicolumn{11}{|c|}{A comparison of challenges among different face datasets} \\
 \hline
 
 Dataset Name & View-point Variation & Different Alignments & Different Expressions \& Emotions & Scaling Changes & Illu-mination Changes & Occlu-sions & Partial Faces & Blur Images & Mirror Reflections & AR Filters\\
 \hline
 
 LFW \cite{lfw} & \textcolor{OliveGreen}{\cmark} & \textcolor{red}{\xmark} & \textcolor{mustard}{\textbf{P}} & \textcolor{red}{\xmark} & \textcolor{mustard}{\textbf{P}} & \textcolor{red}{\xmark} &
 \textcolor{red}{\xmark} &
 \textcolor{red}{\xmark} &
 \textcolor{red}{\xmark} &
 \textcolor{red}{\xmark}\\
 \hline
 
 CASIA WebFace \cite{casiawebface} & \textcolor{OliveGreen}{\cmark} & \textcolor{red}{\xmark} & 
 \textcolor{red}{\xmark} & 
 \textcolor{red}{\xmark} & \textcolor{mustard}{\textbf{P}} & \textcolor{red}{\xmark} &
 \textcolor{red}{\xmark} &
 \textcolor{red}{\xmark} &
 \textcolor{red}{\xmark} &
 \textcolor{red}{\xmark}\\
 \hline
 
VGG Face \cite{vggface} & 
 \textcolor{OliveGreen}{\cmark} & \textcolor{OliveGreen}{\cmark} & \textcolor{OliveGreen}{\cmark} & \textcolor{red}{\xmark} & \textcolor{mustard}{\textbf{P}} & \textcolor{red}{\xmark} &
 \textcolor{mustard}{\textbf{P}} &
 \textcolor{red}{\xmark} &
 \textcolor{red}{\xmark} &
 \textcolor{red}{\xmark}\\
 \hline
 
 UMDFace \cite{umdface} & \textcolor{OliveGreen}{\cmark} & \textcolor{OliveGreen}{\cmark} & \textcolor{OliveGreen}{\cmark} & \textcolor{OliveGreen}{\cmark} & \textcolor{OliveGreen}{\cmark} &
 \textcolor{mustard}{\textbf{P}} &
 \textcolor{mustard}{\textbf{P}} &
 \textcolor{red}{\xmark} &
 \textcolor{red}{\xmark} &
 \textcolor{red}{\xmark}\\
 \hline
 
 WSD & \textcolor{OliveGreen}{\cmark} & \textcolor{OliveGreen}{\cmark} & \textcolor{OliveGreen}{\cmark} & \textcolor{OliveGreen}{\cmark} & \textcolor{OliveGreen}{\cmark} &
 \textcolor{OliveGreen}{\cmark} &
 \textcolor{OliveGreen}{\cmark} &
 \textcolor{OliveGreen}{\cmark} &
 \textcolor{OliveGreen}{\cmark} &
 \textcolor{OliveGreen}{\cmark} \\
\hline
\end{tabular}
\label{tab:challengeComparison}
\end{table*}

\begin{table*}[!t]
\caption{A comparison of the proposed WSD dataset with existing face datasets.}
\centering
\begin{tabular}{ | m{0.17\textwidth} | m{0.09\textwidth} | m{0.08\textwidth} | m{0.09\textwidth} | m{0.45\textwidth} |}
 \hline
 \multicolumn{5}{|c|}{Dataset Statistics Comparison Table} \\
 \hline
 Dataset Name & Number of Subjects & Number of Images & Availability & Annotation Properties\\ 
 \hline
 LFW \cite{lfw} & 5,749 & 13,233 & Public & Several Annotation Attributes\\
 \hline
 CelebFaces \cite{celebfaces} & 10,177 & 202,599 & Private & 5 Landmarks and 40 Binary Attributes\\
 \hline
 CASIA WebFace \cite{casiawebface} & 10,575 & 494,414 & Public & -\\
 \hline
 VGG Face \cite{vggface} & 2,622 & 2,600,000 & Public & Face bounding boxes and Coarse pose\\
 \hline
 UMDFace \cite{umdface} & 8,501 & 367,920 & Public & Face bounding boxes, 21 keypoints, Gender and 3D pose\\
 \hline
 Proposed WSD & 42 & 45,426 & Public & Selfie images, Class labels, Face bounding boxes, and Gender of subjects\\
\hline
\end{tabular}
\label{tab:datasetComparison}
\end{table*}

\subsubsection{AR Filters}
Augmented Reality (AR) is the technology which superimposes a computer-generated image on the user's view of the real-world. There are many smartphone inbuilt applications that provide AR filters like Snapchat, Instagram, B612, etc. that are used by the contributors of the WSD dataset. The AR filters transform the user's image using a wide range of accessories and combinations of special effects, e.g., the Snapchat application transforms the human face into a dog or zombie using animated tricks. While people play with filters for entertainment, the human face still remains visible, and can be both detected and recognized. This makes the dataset more challenging and represents the images mimicing the selfie clicking behaviour of young population.

\subsubsection{Mirrored Reflections}
In the WSD dataset, there are two kinds of mirrored selfies. The first one is due to the pictures taken while standing in front of a mirror by the rear camera of smartphones. The second one is due to the capturing of images through the front camera of smartphones with the selfie containing face reflections.

\subsubsection{Blurred Images}
There are two kinds of blurred images present in the WSD dataset. While some blurred pictures are clicked with the moving camera, others are captured with blur filters `on' in the smartphone. The faces in the selfies are detectable and easily recognizable by the human beings.

\subsubsection{Partial Faces}
Many selfies present in WSD contain only partial visible faces and do not inlude the complete face. Some selfies contain only the left/right half or only the top/bottom portion of the face. A human glance can be easily detected and recognised in these images. Some images are also tilted with multiple alignments in addition to partial faces, adding the diversity and severe challenges in the WSD dataset.

\subsubsection{Occlusions}
The occlusions and obstructions are present in several images of the WSD dataset. Multiple variations of such obstructions such as a facial regions being covered by hand, pen covering the nose tip, AR filters hiding the mouth region, etc. have been included in WSD.

\subsubsection{Illumination Changes}
The WSD dataset incorporates multiple natural lighting conditions in an unconstrained way. Selfies are clicked during dawn, noon, afternoon, dusk, late evening, and even at night. In some images, even more challenging scenarios such as lighting variation from natural light falling on faces from slits through vents or window panes, are present. Collected images have both natural and filtered artificial lighting present.

\subsubsection{Scaling Variations}
Participators clicked selfies in different poses consisting of different sizes of their faces, by changing the camera positions and angles with respect to their faces. Some are full close-up shots covering the facial region from the eyebrows to the chin, while others are shots from far containing the full face along with high degree of background regions.

\subsubsection{Different Expressions and Emotions}
The images in the WSD dataset contain different expressions and emotions of the subjets. Some facial sentiments present include sad, angry, happy, sleepy, excited, irritated, pouting, sulking, crying, smiling, fearful, laughing, yawning, winking, disgusting, and surprised.

\subsubsection{Different Poses and Alignments}
The geometric alignment of selfies also differ greatly in the WSD dataset. As mentioned above, Roll angles in the dataset are diversely varied with no image being strictly vertically aligned leading to different poses and alignments in the WSD dataset.

\subsubsection{View Point Variation}
Multiple profiles of each candidate are included in the WSD dataset as analyzed above in terms of the Yaw characteristics which represent the variations in the view point variations.

\subsubsection{Different Aspect Ratios}
Aspect ratio is the ratio of the width to the height of an image. Since all candidates used the front or back camera of their smartphones and/or mobile devices to click the selfies and videos, the aspect ratios obtained are diverse. While data collection, no restriction on aspect ratio is kept. Some common aspect ratios in the WSD dataset are 1:1, 5:4, 4:3, 3:2, and 16:9. We use the dataset after resizing the images to a fixed size.

\subsection{Comparison with Existing Datasets}
Table \ref{tab:challengeComparison} compares the proposed WSD dataset with the existing face datasets, on the basis of the various challenges present in WSD. We can observe from the existing face datasets that though some challenges like profile variation are present in all the existing datasets, few challenges like scaling, occlusion and partial faces together are present only in a few datasets. The proposed WSD dataset is the only dataset to contain selfies with blurred images, mirrored reflections and AR filters. All the challenges present in the proposed dataset make it highly relevant depicting the real-world scenario. Table \ref{tab:datasetComparison} presents a statistical comparison of the proposed WSD dataset with the existing face datasets. Though the number of subjects and samples are less in the proposed dataset, the images in WSD are collected by the subjects themselves in an unconstrained enviorenement. Whereas the existing datasets are mostly crawled from internet. The existing face datasets do not contain the selfie images which pose its own challenges. Moreover, the proposed dataset contains only the selfie images which have become poupular among young population to capture the images. Further, as mentioned earlier, there exist no publicly available selfie dataset for face detection and recognition in selfie images.

\section{Methodology and Experimental Setup}
In this paper, we adopt the standard face detection (i.e., face region localization) and face recognition methodology for the experiments. Specifically, we consider the existing deep learning based face detection and face recogniton approaches to judge their performance on the proposed WSD dataset. The results of two face detection models and three face recognition models are computed by training the models using the WSD training set and evaluating using the WSD test set. All the networks are trained/fine-tuned for about 100 epochs on our WSD training set using Tesla T4 GPU and evaluated on our WSD test set.

Before using the WSD for face recognition task, we perform face detection using WSD dataset. The state-of-the-art You Only Look Once - v3 (YOLOv3) \cite{yolov3} and Multi-task Cascade Convolutional Neural Network (MTCNN) \cite{mtcnn} models are used to test the face detection performance on the proposed WSD dataset. The YOLOv3 model \cite{yolov3} is a real-time object detection CNN model by utilizing a fast Darknet module as compared to previous versions of YOLO. The MTCNN model \cite{mtcnn} consists of three separate convolutional networks, including P-net, R-net and O-net. The MTCNN model first resizes the input to create the image pyramid followed by non-maximum suppression and bounding box regression for every scaled image to find the face regions in the image. In both cases of YOLOv3 and MTCNN, we use the pretrained models and fine-tune it using our WSD training set and evaluate the model performance on the WSD test set in terms of the mean average precision (mAP). For YOLOv3, we use batch size of 16, learning rate of 0.0001, and batch normalization after each convolution layer for regularization. In case of MTCNN model, we use batch size of 32, learning rate of 0.001, and batch normalization after each convolution layer for regularization.

\begin{figure*}[t]
    \centering
    \includegraphics[width=\textwidth]{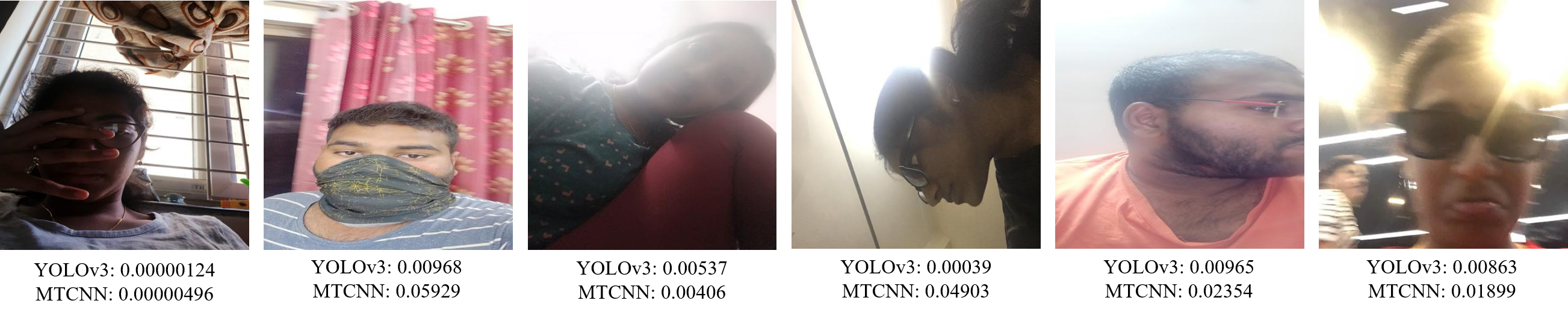}
    \caption{The sample face images for which both the YOLOv3 and MTCNN fail to detect the faces. The score for the corresponding image to have a face using both the models is mentioned.}
    \label{fig:detection_failed}
\end{figure*}

Next we perform the experiments for face recognition on the proposed selfie dataset with three state-of-the-art face recognition models, namely VGGFace \cite{vggface}, VGGFace2 \cite{vggface2} and FaceNet \cite{facenet}. The VGGFace model \cite{vggface} uses CNN descriptors for face recognition. The CNN descriptors are obtained using the VGG-Very-Deep-16 CNN architecture. The VGGFace model is trained on VGGFace dataset consisting of 2,622 identities over 2.6 million images. We use the pretrained VGGFace model and fine-tune it using the WSD training set to evaluate the performance of VGGFace on the WSD test set. In case of VGGFace model, training/fine-tuning on the WSD dataset is done by stochastic gradient descent with 0.9 momentum coefficient using mini-batches of 64 samples and learning rate of 0.01. The VGGFace model is regularised using dropout and weight decay factor of $5 \times 10^-4$. The VGGFace2 model \cite{vggface2} also uses the CNN descriptors for face recognition similar to VGGFace, however, the CNN descriptors are computed using ResNet-50 model \cite{resnet} rather than VGG CNN architecture. The VGGFace2 model is trained on VGGFace2 dataset that consists of 9,131 identities and 3.31 million images. We use the pretrained VGGFace2 model and fine-tune it using the WSD training set to evaluate its performance on the WSD test set. The training configuration of VGGFace2 model is similar to VGGFace, except the Adamax optimizer with learning rate of 0.001 is used for VGGFace2 model. FaceNet model \cite{facenet} uses face embeddings generated using deep CNN like Inception network followed by Triplet loss layer for face recognition and verification. We evaluate the performance of the FaceNet model on the proposed WSD test set by fine tuning the face embeddings of FaceNet model using the WSD training set with triplet loss. In case of FaceNet model, a batch size of 16, learning rate of 0.001 and L2 Regularization are used while fine-tuning face embeddings.

\section{Experimental Results}
In this section, we provide the results obtained using the state-of-the-art deep learning models for face detection and face recognition on the proposed dataset. We also make a comparison of WSD with the benchmark face datasets. 

\begin{table}[t]
\caption{Face detection results (in \% of accuracy) using YOLOv3 \cite{yolov3} and MTCNN \cite{mtcnn} models on WSD dataset in terms of mean average precision (mAP) metric. The existing results on FDDB \cite{fddb} and WIDER FACE \cite{widerface} datasets are also included for a comparison.}
\centering
\begin{tabular}{| p{0.474\columnwidth} | p{0.18\columnwidth} | p{0.18\columnwidth} |}
 \hline
 \textbf{Dataset} & \textbf{YOLOv3} & \textbf{MTCNN} \\\hline
 FDDB \cite{fddb} & 93.90 \cite{li2020face} & 90.20 \cite{li2020face}\\\hline
 Wider Face (Easy) \cite{widerface} & 87.60 \cite{tuli2020novel} & 85.10 \cite{mtcnn}\\\hline
 Wider Face (Medium) \cite{widerface} & 85.80 \cite{tuli2020novel} & 82.00 \cite{mtcnn}\\\hline
 Wider Face (Hard) \cite{widerface} & 72.90 \cite{tuli2020novel} & 60.70 \cite{mtcnn}\\\hline
 \textbf{WSD (Ours)} & 96.92 & 95.77\\\hline
 \end{tabular}
\label{tab:detectionResults}
\end{table}

\subsection{Face Detection Results}
Table \ref{tab:detectionResults} shows the face detection results on WSD and other face datasets, in terms of mAP by using the YOLOv3 \cite{yolov3} and MTCNN \cite{mtcnn} CNN models. The results on the WSD dataset are depicted in the first row. The publicaly available results on benchmark face detection datasets, such as ace Detection Data Set and Benchmark (FDDB) \cite{fddb} and WIDER FACE \cite{widerface} are also included in the table for a comparison. It can be noticed that the performance of both the models on the WSD dataset is higher as compared to FDDB and WIDER FACE dataset. This is due to the nature of images in the proposed and existing datasets. Specifically, the selfie images are generally captured from a camera placed within a limited distance from the subject. This leads to significantly higher facial regions in selfie images as compared to the non-selfie images. Moreover, the majority of the images in the proposed WSD dataset consist of images with single face. Hence, the face detection in the proposed dataset is easier as compared to large-scale non-selfie face datasets. However, the real challenges in the proposed dataset are observed for face recognition as described next.

\begin{figure*}[!t]
    \centering
    \includegraphics[width=\textwidth]{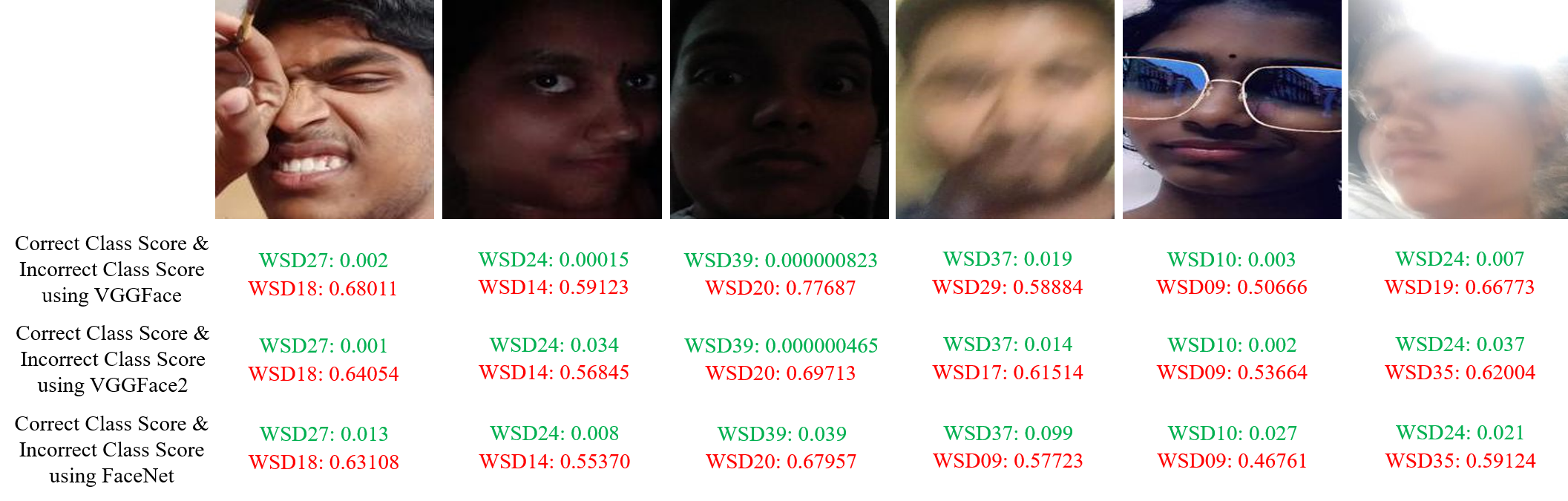}
    \caption{Few sample of failed cases, where all the models misclassify the face to some other person. The correct class with score and incorrect class with score are mentioned in Green and Red color, respectively.}
    \label{fig:classification_failed}
\end{figure*}

We report some failure cases where both the YOLOv3 and MTCNN models are not able to detect the face regions in Fig. \ref{fig:detection_failed} alongwith the score of having the face. It can be seen that the score to detect a face in the images is very low using both the models, which indicate the complexity of the images for face detection. Upon analysis with the failure cases, we notice that the selfies with severe background illumination changes are not being detected, irrespective of whether the face is clearly visible or challenging. In addition to this, people having occlusions near the eye are harder to detect. Partial images are another diffuculty to detect and in many cases, images containing AR filters had incorrect detections. In such cases bounding box does not contain the face, but other regions. These are the challenges for the face detection task in selfie images.

\begin{table}[!t]
\caption{Face recognition results (in \% of accuracy) using VGGFace, VGGFace2 and FaceNet models on WSD dataset in terms of validation accuracy metric. The existing results on LFW \cite{lfw}, CASIA WebFace \cite{casiawebface}, VGGFace2 \cite{vggface2}, YouTube Faces \cite{wolf2011face} and IJB-A \cite{ijba}, IJB-B \cite{ijbb}, and IJB-C \cite{ijbc} face datasets are also included for a comparison.}
\centering
\resizebox{\columnwidth}{!}{
\begin{tabular}{| p{0.307\columnwidth} | p{0.18\columnwidth} | p{0.18\columnwidth} | p{0.26\columnwidth} |}
\hline
Dataset & VGGFace & VGGFace2 & FaceNet \\\hline
LFW & 98.95 \cite{vggface} & - & 99.63 \cite{facenet}\\\hline
CASIA WebFace & 90.7 \cite{zhang2017local} & - & 99.05 \cite{william2019face}, \cite{xu2020lightweight}\\\hline
VGGFace2 & 89.4 \cite{vggface2} & 96.1 \cite{vggface2} & 99.65 \cite{william2019face}, \cite{xu2020lightweight}\\\hline
YouTube Faces & 97.3 \cite{vggface} & - & 95.12 \cite{facenet} \\\hline
IJB-A & 95.4 \cite{vggface2} & 98.0 \cite{vggface2} & - \\\hline
IJB-B & 85.0 \cite{vggface2} & 93.8 \cite{vggface2} & - \\\hline
IJB-C & - & 95.0 \cite{vggface2} & - \\\hline
\textbf{WSD (Ours)} & 88.53 & 92.78 & 93.98\\\hline
\end{tabular}
}
\label{tab:recognitionresults}
\end{table}

\subsection{Face Recognition Results}
Table \ref{tab:recognitionresults} shows the results of applying VGGFace \cite{vggface}, VGGFace2 \cite{vggface2} and FaceNet \cite{facenet} face recognition techniques on different face reginition datasets including the WSD. The first row depicts the face recognition results on the proposed WSD dataset. In order to perform the results comparison, the existing results on the benchmark datasets, such as LFW \cite{lfw}, CASIA WebFace \cite{casiawebface}, VGGFace2 \cite{vggface2}, YouTube Faces \cite{wolf2011face}, Iarpa Janus Benchmark-A (IJB-A) \cite{ijba}, IJB-B \cite{ijbb} and IJB-C \cite{ijbc}, are also reported. It is quite clear from the face recognition results that the performance of the face recognition models is significantly lower on the proposed WSD dataset as compared to the existing face datasets. The lower face recognition performance on the proposed WSD dataset is due to the inherent challenges present in the dataset, such as illumination changes, AR filters, occlusion, scale changes, blur, pose variations, and many more. 

Sample images are shown in Fig. \ref{fig:classification_failed} where all the classification models fail to classify the faces into correct class. The correct class alongwith score for each model is shown in Green color. Whereas, the incorrectly recognized classes alongwith score for each model is shown in Red color. It can be seen that the class score for correct class is very low for all the models. On the other hand, the class score for incorrect class is very high. This shows the complexity level of the images present in the developed dataset. Overall, the experimental results confirm that the proposed WSD dataset is challenging for face recognition in selfie images.

\section{Conclusion and Future Directions}
A wild selfie dataset (WSD) is proposed in this paper, for face recognition in selfie images. The proposed dataset includes the face images of mostly youngsters with more female subjects than male subjects. The WSD dataset is very challenging in several terms, such as images containing effects of AR filters, mirrored reflections, blur, partial faces, occlusions, illumination changes, scaling, different expressions and emotions, different alignments, view point variation and different aspect ratios. The proposed WSD dataset is tested for face detection and face recogntion tasks. The face detection results are computed using YOLOv3 and MTCNN models. It is found that the face detection results are satisfactory on the WSD dataset as the selfie images are captured from a limited distance. However, the presence of severe illumination changes and occlusions pose significant challenges to the face detection models. The face recognition results are computed using VGGFace, VGGFace2 and FaceNet models. We notice that the performance of the face recogntion models is significantly lower as compared to the existing face datasets. The face recognition results confirm the complexity of the proposed WSD dataset, consisting of real challenges. The proposed WSD dataset can significantly boost the advancements in the face recognotion technology to better tackle the challenges posed while capturing the selfie images in real scenario.

{\small
\bibliographystyle{ieee_fullname}
\bibliography{Reference}
}

\end{document}